\documentclass[twocolumn,letterpaper]{IEEEAerospaceCLS}  % only supports two-column, letterpaper format

\usepackage[]{graphicx}    % We use this package in this document
\newcommand{\ignore}[1]{}  % {} empty inside = %% comment
 \usepackage{url}
\usepackage{amsmath,amssymb,amsfonts}
\usepackage{algorithmic}
\usepackage{graphicx}
\usepackage{textcomp}
\usepackage{xcolor}
\usepackage{multirow}
\usepackage{gensymb}
\usepackage{float}
\usepackage{makecell}
\usepackage{inputenc}
\DeclareUnicodeCharacter{2212}{-}
\usepackage{multirow}
\usepackage{algorithmic}
\usepackage{graphicx}
\usepackage{textcomp}
\usepackage{xcolor}
\begin{document}
\title{Sliding Window Neural Generated Tracking Based on Measurement Model}

\author{%
Haya Ejjawi\\ 
Sorbonne University-Abu Dhabi\\Department of Science and Engineering\\
Abu Dhabi, UAE\\
A00021335@sorbonne.ae
\and 
Amal El Fallah Seghrouchni\\
Mohammed VI Polytechnic University\\Ai movement\\
Rabat, Morocco\\
Sorbonne University\\
Paris, France\\
Amal.Elfallah@lip6.fr
\and
Frederic Barbaresco\\
Thales Group\\
Paris, France\\
frederic.barbaresco@thalesgroup.com
\and
Raed Abu Zitar\\
Thales/SCAI Senior Scientist\\
Sorbonne University-Abu Dhabi\\
Abu Dhabi, UAE\\
raed.zitar@sorbonne.ae
%%%% IMPORTANT: Use the correct copyright information--IEEE, Crown, or U.S. government. %%%%%
\thanks{\footnotesize 978-1-6654-9032-0/23/$\$31.00$ \copyright2023 IEEE}              % This creates the copyright info that is the correct 2023 data.
%\thanks{{U.S. Government work not protected by U.S. copyright}}         % Use this copyright notice only if you are employed by the U.S. Government.
%\thanks{{978-1-6654-9032-0/23/$\$31.00$ \copyright2023 Crown}}          % Use this copyright notice only if you are employed by a crown government (e.g., Canada, UK, Australia).
%\thanks{{978-1-6654-9032-0/23/$\$31.00$ \copyright2023 European Union}}    % Use this copyright notice is you are employed by the European Union.
}

\maketitle

\thispagestyle{plain}
\pagestyle{plain}

\maketitle

\thispagestyle{plain}
\pagestyle{plain}

\begin{abstract}
In the pursuit of further advancement in the field
of target tracking, this paper explores the efficacy of a feedforward neural network in predicting drones’ tracks, aiming to
eventually, compare the tracks created by the well-known Kalman
filter and the ones created by our proposed neural network.
The unique feature of our proposed neural network tracker is
that it is using only a measurement model to estimate the next
states of the track. Object model selection and linearization is
one of the challenges that always face in the tracking process.
The neural network uses a sliding window to incorporate the
history of measurements when applying estimations of the track
values. The testing results are comparable to the ones generated
by the Kalman filter, especially for the cases where there is
low measurement covariance. The complexity of linearization is
avoided when using this proposed model.
\end{abstract}

\tableofcontents

\section{Introduction}
\noindent Many variations of the Kalman filter have been derived with each being the most convenient according to the problem at hand. The most well-known types of Kalman filters are the linear Kalman filter, the Extended Kalman filter, and the Unscented Kalman filter\cite{welch1995introduction}. Linear Kalman filter is a recursive algorithm that fuses measurements from several sensors to estimate in real time the state of a robotic system when the estimation system is linear and Gaussian. An estimation system is considered linear when both the motion model and the measurements model are linear \cite{rhudy2017kalman}. The state equation can be generalized to:\\
%\vspace{0.2cm}
%\begin{equation}
\centerline{$x_{k+1} = A_{k}x_{k}+B_{k}u_{k}+G_{k}v_{k}$}
%\end{equation}
%\vspace{0.2cm}
\newline where $A_{k}$ is the state transition matrix and $B_{k}$ is the control matrix, which accounts for any unknown forces acting on the object. $v_{k}$ represents process noise following a Gaussian distribution of mean 0 and covariance $Q_{k}$. $G_{k}$ is the process noise gain matrix.
\vspace{0.2cm}
\newline Measurements are what is observed or measured in a system. A linear measurement can be represented as:
\vspace{0.2cm}
\centerline{$z_{k} = H_{k}x_{k}+w_{k}$}
%\vspace{0.2cm}
\newline Where $H_{k}$ is the measurement matrix and $w_{k}$ represents measurement noise at the current time step. The measurement noise is Gaussian and of zero means. The covariance matrix of the measurement noise is assumed to be $R_{k}$. The filter works in two steps, prediction, and correction. In the prediction part, the system model is used to calculate the \textit{a priori} state $x_{k+1 \mid k}$ and its covariance matrix $P_{k+1 \mid k}$, given initial estimates $x_{0\mid0}$ and $P_{0\mid0}$. The subscript $k+1\mid k$ indicates that the corresponding quantity is the estimate of the k+1 step propagated from the $k^{th}$ step. In the second step the filter uses the a priori estimates calculated in the first part to compute the \textit{posteriori} estimates $x_{k+1 \mid k+1}$ and $p_{k+1 \mid k+1}$. The process can be summarized in the table below: 
\vspace{0.2cm}
\begin{table}[H]
\caption{Kalman Filter Algorithm}
\scalebox{0.7}{
\centering
\begin{tabular}{|c|c|} 
 \hline
   Prediction & Correction \\ 
   %[0.5ex] 
 \hline\hline
%\makecell
$x_{k+1\mid k}=A_{k}x_{k\mid k}+B_{k}u_{k}$ & $S_{k+1}=H_{k+1}P_{k+1\mid k}H_{k+1}^{T}+R_{k+1}$ \\
 \vspace{0.1cm}
 $P_{k+1\mid k}=F_{k}P_{k\mid k}F_{k}^{T}+G_{k}Q_{k}G_{k}^{T}$ & $K_{k+1}=P_{k+1\mid k}H_{k+1}^{T}S_{k+1}^{-1}$\\
 $z_{k+1\mid k}=H_{k+1}x_{k+1\mid k}$  &                       $x_{k+1\mid k+1}=x_{k+1\mid k}+K_{k+1}(z_{k+1}−z_{k+1\mid k})$ \\
   & $P_{k+1\mid k+1}=P_{k+1\mid k}− K_{k+1}S_{k+1}K_{k+1}^{T}$ \\ 
 \hline
\end{tabular}}
\label{table:1}

\end{table}
\vspace{0.2cm}
\noindent Where $K_{k+1}$ is the Kalman gain, $S_{k+1}$ is the measurement predicted covariance matrix, and $x_{k+1 \mid k+1}$ is the corrected state at step k+1, and $P_{k+1\mid k+1}$ is its corrected covariance matrix at step k+1.
\vspace{0.2cm}
\newline
\noindent Where $K_{k+1}$ is the Kalman gain, $S_{k+1}$ is the measurement predicted covariance matrix, and $x_{k+1 \mid k+1}$ is the corrected state at step k+1, and $P_{k+1\mid k+1}$ is its corrected covariance matrix at step k+1.
\vspace{0.2cm}
\newline
The Extended Kalman filter (EKF) is the nonlinear version of the Kalman filter \cite{terejanu2008extended}. EKF is used in this work and it works on two major assumptions, the system is linear and Gaussian \cite{laaraiedh2012implementation}. In order to use the Kalman filter with nonlinear functions, linearization is required. The problem that arises when dealing with nonlinear functions is that feeding a Gaussian distribution with a nonlinear function results in a non-Gaussian distribution and therefore Kalman filter cannot be used. To overcome this issue, a linear approximation of nonlinear functions is needed and can be achieved with the help of the Taylor series\\
%begin{equation}
%\left( x \right) & = \sum\limits_{n = 0}^\infty %{\frac{{{f^{\left( n \right)}}\left( a \right)}}\\{{\left( {x - a} \right)}^n}} \\ & = f\left( a \right) +\\ f'\left( a \right)\left( {x - a} \right) + \\\frac{{f''\left( a \right)}}{{2!}}{\left( {x - a} \right)^2} +\\ \frac{{f'''\left( a \right)}}{{3!}}{\left( {x - a} \right)^3} + \cdots \end{equation}
\noindent for linearizing, only first-order approximation is required. The approximation of the nonlinear curve is done at the mean of the Gaussian distribution. In this problem, the extended Kalman filter is used since measurements are collected from a simulated 2-D bearing range sensor as polar coordinates. 
\vspace{0.2cm}
\newline
On the other hand, a feed-forward neural network will be used to be trained with the measurements data \cite{sandberg2001nonlinear}, \cite{priddy2005artificial}, \cite{hassoun1995fundamentals}. Feedforward neural networks are artificial neural networks where the connections between units do not form a cycle. In a feed-forward neural network, information only travels forward in the network, first through the input nodes, then through the hidden layers, and finally through the output nodes. Backpropagation algorithm is used to find the optimal parameters \cite{rumelhart1986learning}.
The paper is organized as follows; section 3 will cover the methodology, section 4 will cover simulations and results, and section 5 will summarize the conclusion and results.

%%What is sensor fusion algorithm? Sensor fusion algorithms combine sensory data that, when properly synthesized, help reduce uncertainty in machine perception. They take on the task of combining data from multiple sensors — each with unique pros and cons — to determine the most accurate positions of objects
\section{ Methodology of Training and Testing}
Stone soup, a software that provides a framework for target tracking and state estimation and enables the development and testing of state estimation algorithms, is first used to create a target that moves with a near-constant velocity from the (0,0) location point. \cite{barr2022stone}. Measurements are then obtained through a 2-D bearing and range sensors. Measurements are given as polar coordinates so conversion to Cartesian coordinates is required:
\vspace{0.2cm}
\newline
\begin{center}
$\begin{bmatrix} x \\ y \end{bmatrix}$ = $\begin{bmatrix} rcos(\theta)\\ rsin(\theta) \end{bmatrix}$
\end{center}
\vspace{0.2cm}
%\newline
However, the sensor offset must be taken into account when converting. In this case, the sensor is placed at (50,0). For obtaining the correct x-coordinate, 50 must be added to the X-location value. The EKF is then used to obtain a track. Several data sets are generated by setting different  seeds and based on different given ground truth tracks.\\

\noindent For initial training and testing \cite{sazli2006brief}, data is divided into training and testing sets for better initial evaluation of the model. The model we are proposing is called the sliding window method. This method is implemented such that $n$ measurements  are fed into the neural network (NN)  at a time $t$. For example, if $n=3$ (number of measurement values),  then measurement values at time $t$, $t-1$, and $t-2$ are fed simultaneously to the neural network. The neural network calculates the predicted output $\hat{Y}_{t}$.  The given ground truth value $Y_{t}$ is the target value at $t$. Please, see Figure 1 for more clarifications.  The error that is used in adjusting the weights of the neural net is evaluated by the Mean Squared Error (MSE). This measure is based on the sum of the euclidean distances between the ground truth values and the neural network output values for a sequence of $m$ length, (see equations 1 and 2):\\
\begin{equation}
    MSE=\frac{1}{m}\sum_{i=1}^{m}(Y_{i}-\hat{Y}_{i})^{2}\\
\end{equation}\\

\noindent MSE	=	mean squared error\\
{n}	=	number of data points\\
$Y_{i}$	=	observed values (neural network targets)\\
$\hat{Y}_{i}$	=	predicted values (neural network outputs\\
\begin{equation}
    d(p,q)=\sqrt{\sum_{i=1}^{n}(q_{i}-p_{i})^{2}}
\end{equation}\\
\noindent 
p,q	=	two points in Euclidean n-space\\
$q_i, p_i$	= $i^{th}$ coordinate of the points p \& q	\\
n	=  dimension of the space

\vspace{0.2cm}

\noindent First, weights are initialized randomly in order to start the forward propagation process. In forward propagation, an input feature vector is initially multiplied with a weight matrix and a bias vector is added.
Then, in order to achieve non-linearity, the weighted sum is fed into an activation function \cite{nwankpa2018activation}, \cite{ramachandran2017searching}. The nonlinear Relu Activation function $f(Z)$ is used for all units of layers [l]; l < L:\\
\vspace{0.1cm}
\begin{equation}
  f(Z)=max(0,Z)
\end{equation}

and its first order derivative:\\
\vspace{0.1cm}
\begin{equation}
f'(Z)=
\begin{cases} 
0 & \text{if  }  Z < 0 \\
1 & \text{if  }  Z > 0 \\
\end{cases}
\end{equation}

\vspace{0.1cm}
\noindent This provides the output of the $j^{th}$ unit of the layer. Multiplications are simultaneously performed by multiple units of layer [l] providing an output vector of length equal to the number of units. This output vector is then multiplied by the weight matrix of the subsequent layer [l+1] and fed into the activation function of layer [l+1]. After repeating this process L-1 times (for an L-Layer Neural Network), the output vector (at layer L-1) is fed into a final activation function. In this case, the final activation function used is the identity in order to obtain a real-numbered continuous outcome. The cost function measures how well the parameters perform on the training set \cite{krogh2008artificial}. In this case, the aim is to minimize the sum of least squares, also known as the Ordinary Least Squares method (OLS):\\
\begin{equation}
L(a,y) = \frac{1}{m}\sum_{i=1}^{m} \ (a_i - y_i)^2 \,    
\end{equation}
\newline
\vspace{0.1cm}
\noindent The back-propagation computes the gradient  of the loss function with respect to the weights \cite{yu2002general}. The gradients are computed iteratively starting at the final Layer L \cite{nielsen2015neural}. The algorithm used in this problem for getting the best model parameters is called Gradient Descent.
\noindent The general work procedure can be summarized in the following flow chart:\\
\begin{figure}[H] 
\scalebox{0.6}{
\centering   
\includegraphics{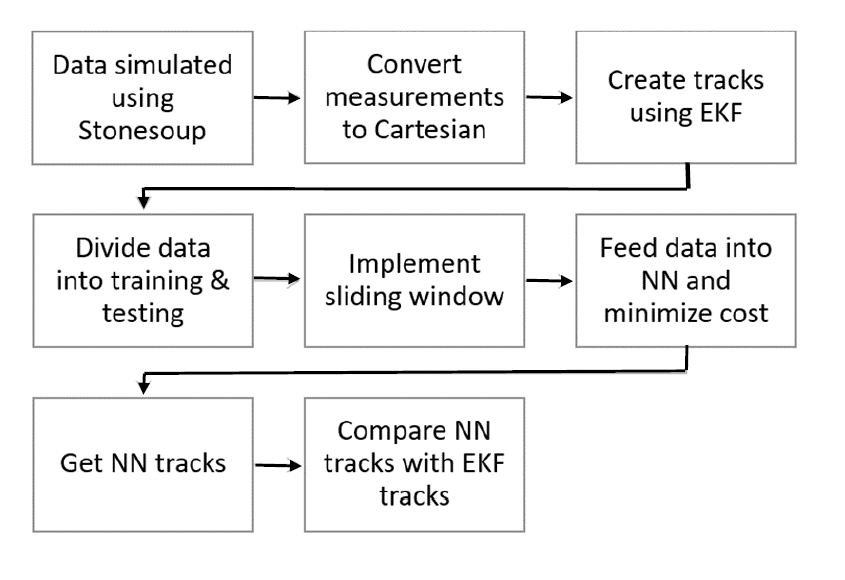}}
\caption{Flow of Tracks Generation.}
\label{fig:flow chart}
\end{figure}

\begin{figure}[H]
\scalebox{0.65}{
\centering
\includegraphics{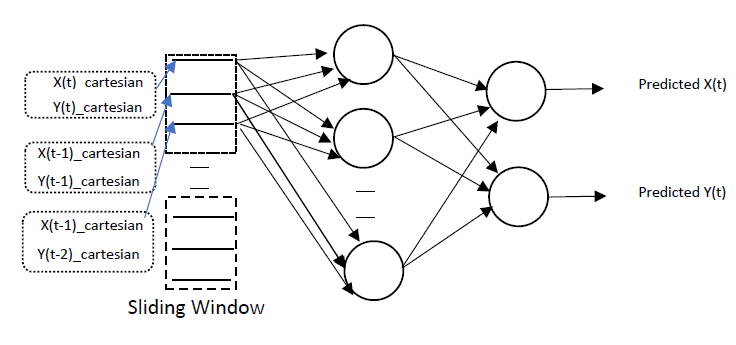}}
\caption{Sliding Window Neural Network}
\label{NN_Sliding window}
\end{figure}

%\vspace{0.3cm}

%\vspace{0.2cm}

%\section{Neural Network tracking with Sliding Window}
The sliding window neural network, as shown in Figure 2,  implicitly includes a short history of the measurement values,in addition to the current input (for example $t$, $t-1$, $t-2$ for $n=3$), and use them as the input vector. Then it slides up one time step and uses the measurements at $t+1$, $t$, and $t-1$ as a new vector of measurement inputs and the new target will be the ground truth at $t+1$. It keeps doing this for a predefined sequence of times $m$ during the training phase.  During the testing phase, the process is repeated but with no adjustment for weights and using data that never was used in training. Keep in mind that every input and the target value is a 2-tuple vector of $X_{Cartesian}$ and $Y_{Cartesian}$ location values. Originally Stone Soup software provides the range and the bearing values, as mentioned earlier.   
%%%%%%%%%%%%%%%%%%%%%%%%%%to here Sunday Sept-25
\section{ Simulations and Results: KF and NN}
Below is depicted the flow of simulations and testing processes shown in the flowchart of Figure .2. The process uses both the neural network and the Extended Kalman filter (EKF) model that is provided by stone soup open source software 
\url{ https://stonesoup.readthedocs.io/en/v0.1b5/} in order to generate input measurements and EKF tracking results. The tracked variables are only the locations (converted to $X_{cartesian}$ and $Y_{cartesian}$) for  drone (flying object) and in 2-D.
\subsection{Training \& Testing}
Through the sliding window method, the number of sequences used in training and testing is set to  $m=5$ so data is read such that every 3 measurement points ($n=3)$ (a point has two tuples of X and Y) is mapped to one ground truth point (another X and Y tuple). You can say that memory of length $3$ is implicitly used here, see Figure 1, please. Initially, the first 4 rows are used for training and the $5th$ is kept for testing and not used in training. Each row contains 3 tuples for training and the last tuple as a target (ground truth). The number of layers, number of nodes, and the learning rate are chosen empirically. Weights are initially set randomly from a standard normal distribution and the biases are set to zero. Cost is traced every 100 iterations.
%\vspace{0.2cm}
\begin{table}[H]
\caption{ Five rows of the training data (each row is 3 Measurements + 1 Ground truth)}
\scalebox{0.7}{
\begin{tabular}{|l|l|l|l|l|l|l|l|}
\hline
X(M).1   & Y(M).1   & X(M).2   & Y(M).2   & X(M).3   & Y(M).3   & X(G)   & Y(G)   \\
\hline
2.2218 & -0.95  & 1.549  & -2.188 & 4.1186 & 11.127 & 2.2755 & 2.0307 \\
1.549  & -2.188 & 4.1186 & 11.127 & 4.5342 & 4.7703 & 3.156  & 3.2077 \\
4.1186 & 11.127 & 4.5342 & 4.7703 & 4.473  & 4.422  & 4.1115 & 4.6815 \\
4.5342 & 4.7703 & 4.473  & 4.422  & 5.8321 & 9.1323 & 5.0689 & 6.4241 \\
4.473  & 4.422  & 5.8321 & 9.1323 & 4.6718 & 9.0161 & 6.1341 & 8.2916 \\
\hline
\end{tabular}}

\label{table:2}
\end{table}
\vspace{0.2cm}
\noindent The inputs of the training model are:
\begin{itemize}
    \item Number of hidden layers: 3
    \item Number of nodes in each hidden layer: 7 \& 5 \& 4
    \item Number of iterations: 20000
    \item Learning rate: 0.0001
\end{itemize}
After 20000 iterations, the cost drops to 0.017( See Figure 3, please):
\begin{figure}[H]
    \centering
     \scalebox{0.8}{
    \includegraphics[width=8cm, height=8cm]{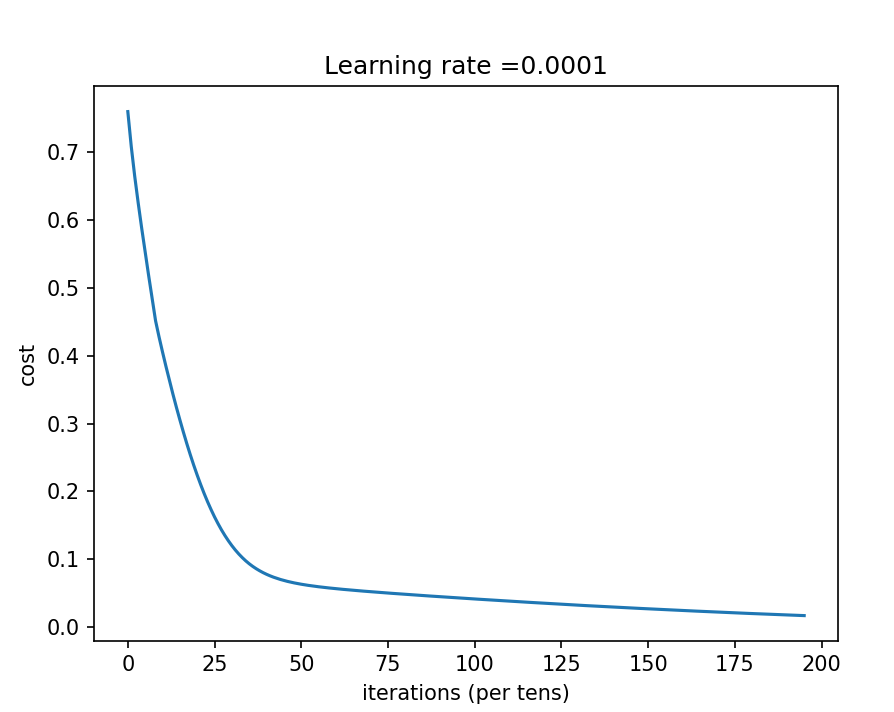}}
    \caption{Cost Plot}
    \label{fig:my_label}
\end{figure}
\noindent The predicted output of the training set is shown in Table III below:
\begin{table}[H]
 \caption{Training Set NN Output (predicted track)}
    \centering
    \begin{tabular}{|c|c|c|c|c|}
    \hline
      x   &  2.18 & 3.036 & 4.09 & 5.19\\
    \hline
      y   & 2.11 & 3.30 & 4.69 & 6.33  \\
    \hline
    \end{tabular}
    \label{tab:figure1}
\end{table}
\noindent Plotting the neural network track obtained from the training set output along with the ground truth and EKF track yields tracks in Figure 4. The testing is implemented on measurement values $$(x,y)$$ that was never used in training. Usually, around 70\% of the measurement values are used in training, and the rest are used in testing :
\begin{figure}[H]
    \centering
     \scalebox{0.8}{
    \includegraphics[width=10cm]{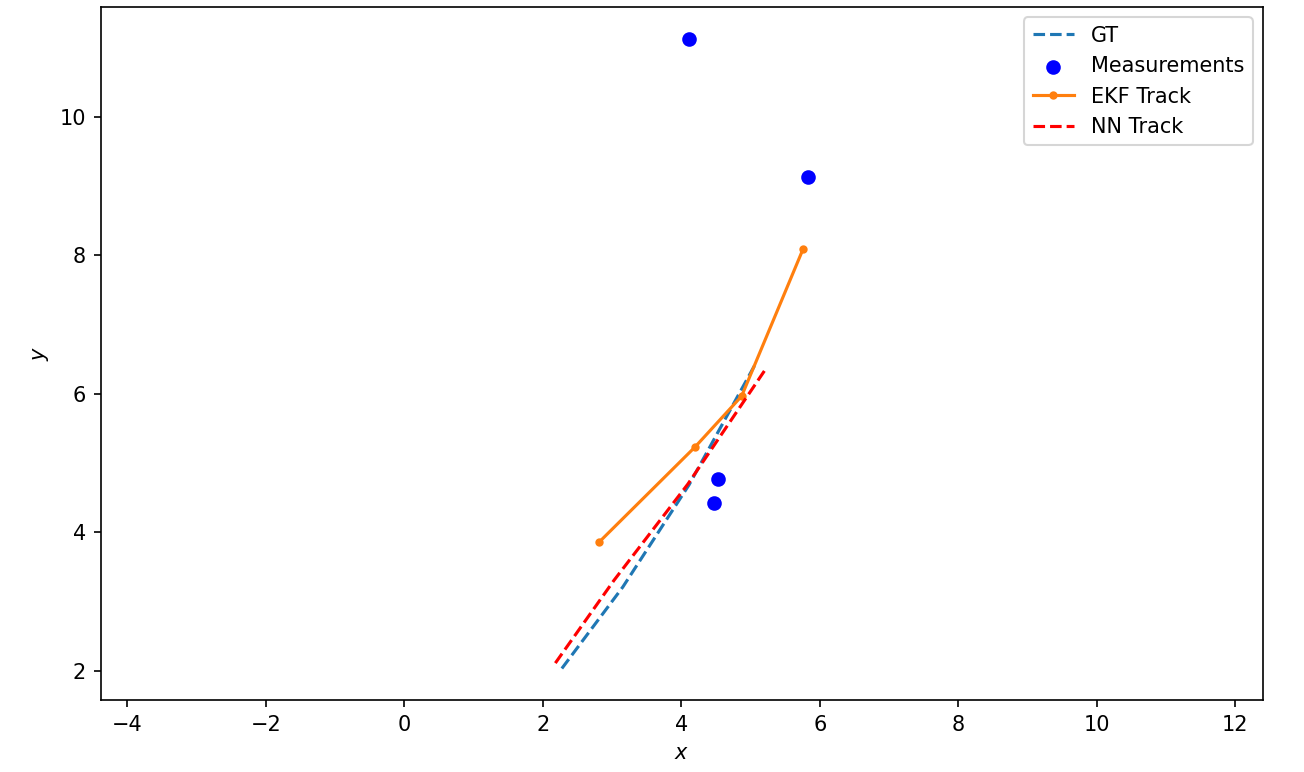}}
    \caption{NN training output, (Ground truth) GT, measurements,\& EKF}
    \label{fig:figure2}
\end{figure}
\noindent The mean squared error for the training set is: 0.01\\
The Euclidean distance between the neural network output for the training set and the ground truth is: 0.465\\
The Euclidean distance between the Extended Kalman Filter track corresponding to the training set (measurements) and the ground truth is: 7.465\\
\vspace{0.2cm}
\newline
\noindent The output of the testing set is shown in TABLE IV:\\
\vspace{0.2cm}

\begin{table}[H]
    \centering
    \caption{Testing Set NN Output after using the fifth row in Table II as input test}
    \begin{tabular}{|c|c|}
    \hline
      x   &  4.96\\
    \hline
      y   & 5.84 \\
    \hline
    \end{tabular}
    \label{tab:figure2}
\end{table}

\noindent The mean squared error for the testing set is: 1.38\\
The Euclidean distance between the neural network output of the testing set and the ground truth is 2.722\\
The Euclidean distance between the Extended Kalman Filter track corresponding to the testing set and the ground truth is 1.364\\

%\vspace{0.3cm}

\noindent The neural network is then used to train 15 rows of data, and 4 rows are kept for testing(see Tables V and VI). The ground truth is provided below for comparison:\\
\vspace{0.2cm}

\begin {table}[H]
\caption{Training Set GT}
\scalebox{0.45}{\begin{tabular}{|l|l|l|l|l|l|l|l|l|l|l|l|l|l|l|l|l|l|l|l|}
\hline
X(G) & 2.28 & 3.16 & 4.11 & 5.07& 6.13 & 7.34 & 8.71 & 9.66 & 10.23 & 10.91 & 11.44 & 12.087 & 12.866 & 13.66 & 14.66 \\
\hline
Y(G) & 2.03 & 3.21 & 4.68 & 6.42 & 8.29 & 10.21 & 12.64 & 15.08 & 17.46 & 19.92 & 22.44 & 25.23 & 28.32 & 31.65 & 34.96 \\
\hline

\end{tabular}}
\label{tab:figure3}
\end{table}
%\vspace{0.2cm}
\begin{table}[H]
\caption{Testing Set GT}\
\centering
\scalebox{0.9}{\begin{tabular}{|l|l|l|l|l|}
\hline
X(G) & 15.76&	16.99&	18.30&	19.72 \\
\hline
Y(G) &  38.33&	41.65&	44.95&	48.07\\
\hline

\end{tabular}}
\label{tab:figure3}
\end{table}
\vspace{0.2cm}
\noindent The inputs of the training model are:\\
\begin{itemize}
    \item Number of hidden layers: 3
    \item Number of nodes in each hidden layer: 7 \& 5 \& 4
    \item Number of iterations: 2000000
    \item Learning rate: 0.0001
\end{itemize}
After 2000000 iterations, the cost drops to 0.097.
%begin{figure}[H]
%\scaleb.ox{09}{
 %   \centering
 %   \includegraphics[width=10cm]{Cost plot cost 0.09.PNG}}
 %   \caption{Cost Plot}
%\label{fig:my_label}
%\end{figure}
\noindent The predicted output of the training set is (Table VII):\\
\begin{table}[H]
\caption{Training Set NN Output}
\scalebox{0.5}{\begin{tabular}{|l|l|l|l|l|l|l|l|l|l|l|l|l|l|l|l|l|l|l|l|}
\hline
X & 2.33 & 3.25 &  4.29 & 5.19 &  5.99 &
         7.21 &  8.49 &  9.38 & 10.79 & 10.55 &
        11.21 & 12.39 & 13.09 & 13.88 & 14.63   \\
\hline
Y & 2.04& 3.2 &4.69 &6.51 &8.19
 &10.21
 &12.49
 &14.82
 &17.47
 &19.84
 &23.03
 &25.21
 &28.88
 &31.68
 &34.29 \\
\hline

\end{tabular}}
\label{tab:figure3}
\end{table}

\noindent Plotting the neural network track obtained from the training set output along with the ground truth and EKF track yields (Figure. 5):\\

\begin{figure}[H]
    \centering
     \scalebox{0.8}{
    \includegraphics[width=10cm]{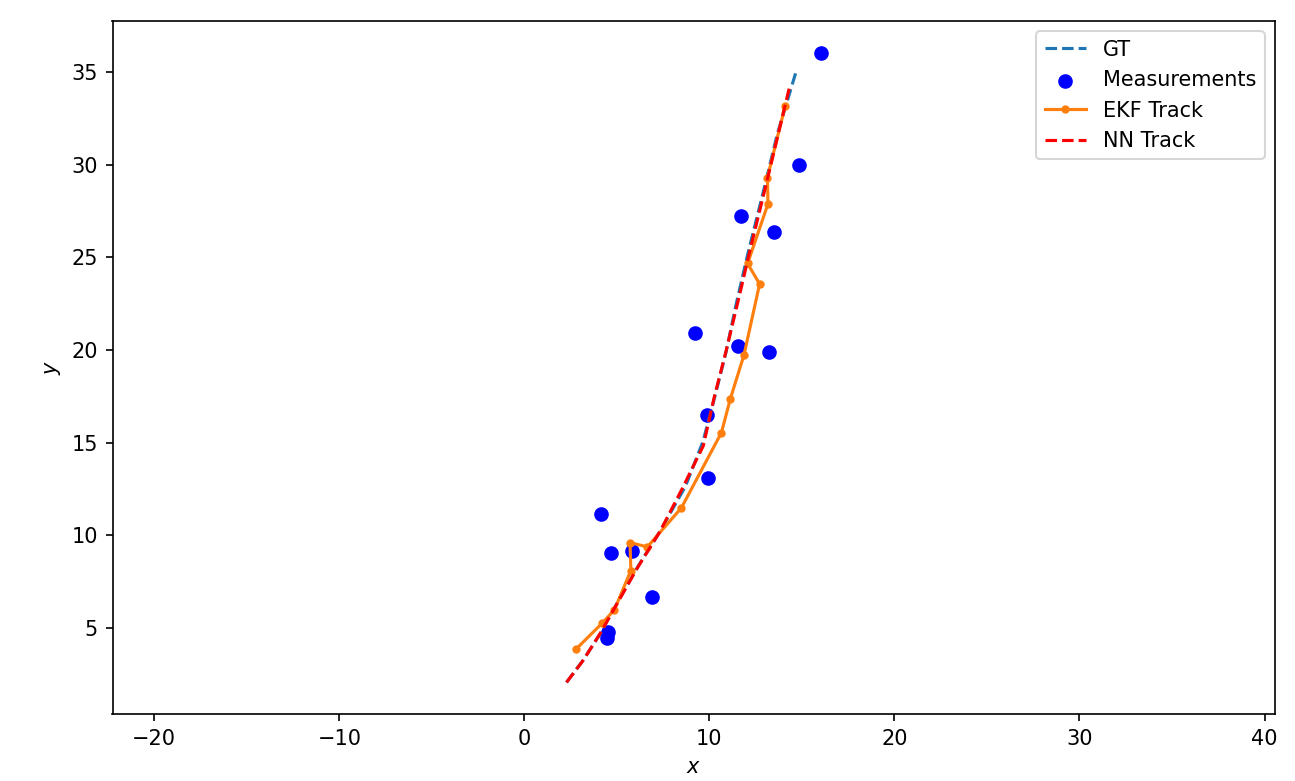}}
    \caption{NN Training Output, GT, \& EKF}
    \label{fig:my_label}
\end{figure}
\noindent The mean squared error for the training set is: 0.015\\
The neural network predictions of the training set are visually and by inspection close to the ground truth values. \newline It is worth informing that in all previous simulations, the default covariance values for the bearing angle and the range have the following values( $\sigma_{11}$ is for the bearing, and $\sigma_{22}$ is for the range):
\newline
%\vspace{2.0cm} 
\centerline{$\begin{bmatrix}
0.00349066 & 0\\
0 & 0.5  
\end{bmatrix}$}\\
%vspace{0.2cm}

%%%%%%%%%%%%%%%%%%%%%%%%%to here Sept 26
\section{Experimenting With Different Noise Covariances}
\noindent Unlike the Kalman filter, the neural network is not provided with the object's motion model, but only with measurements. It is therefore plausible to assume that measurements with lower covariance, thus closer to the ground truth values, would result in better testing outcomes for the neural network. The measurements noise covariance matrix is currently following the stone soap default setting mentioned above. 

%%\begin{figure}[H]
%    \centering
%\scalebox{0.8}{
%    \includegraphics[width=10cm]}
  %  \caption{Measurements \& Ground Truth}
 %   \label{fig:my_label}
%\end{figure}
\subsection{Setting the Covariance to 0.001}
Setting both values on the diagonal of the measurements covariance matrix to 0.001 (Figure 6):\\
\vspace{0.2cm} 
\centerline{$\begin{bmatrix}
0.001 & 0\\
0 &   0.001 
\end{bmatrix}$}
%\vspace{0.2cm}
%\begin{figure}[H]
%\scalebox{0.8}{
  %  \centering
  %%  \includegraphics[width=10cm]{measurements with grounf truth %covar 0.PNG}}
   % \caption{Measurements \& Ground Truth, Covariance 0.001}
   % \label{fig:my_label}
%\end{figure}
\noindent Measurements are now less scattered around the ground truth. Ground truth values remain the same. Data, as previously explained, is divided into 15 rows for training and 4 for testing after implementing the sliding window. Repeating the training process with the following inputs:\\
\begin{itemize}
    \item Number of hidden layers: 3
    \item Number of nodes in each hidden layer: 7 \& 5 \& 4
    \item Number of iterations: 2000000
    \item Learning rate: 0.0001
\end{itemize}
After 2000000 iterations, the cost drops to  0.0102.
%\begin{figure}[H]
%%   \centering
 %   \includegraphics[width=10cm]{cost covarinace 0.PNG}}
 %   \caption{Cost Plot}
%\label{fig:my_label}
%\end{figure}
\noindent The predicted output of the training set is (shown in Table VIII):\\
\begin{table}[H]
\caption{Training Set NN Output}
\scalebox{0.5}{\begin{tabular}{|l|l|l|l|l|l|l|l|l|l|l|l|l|l|l|l|l|l|l|l|}
\hline
X & 2.23&
 3.24&
 4.07&
 5.05&
 6.22&
 7.34&
 8.69&
 9.66&
 10.31&
 10.89&
 11.4&
 11.93&
 12.81&
 13.92&
 14.6   \\
\hline
Y & 2.06&
 3.15&
 4.73&
 6.42&
 8.25&
 10.21&
 12.67&
 15.06&
 17.52&
 19.9&
 22.36&
 25.35&
 28.29&
 31.61&
 34.97 \\
\hline

\end{tabular}}
\label{tab:figure3}
\end{table}

\noindent Plotting the neural network track obtained from the training set output along with the ground truth and EKF track yields (Figure 6):\\
\begin{figure}[H]
    \centering
    \scalebox{0.8}{
    \includegraphics[width=10cm]{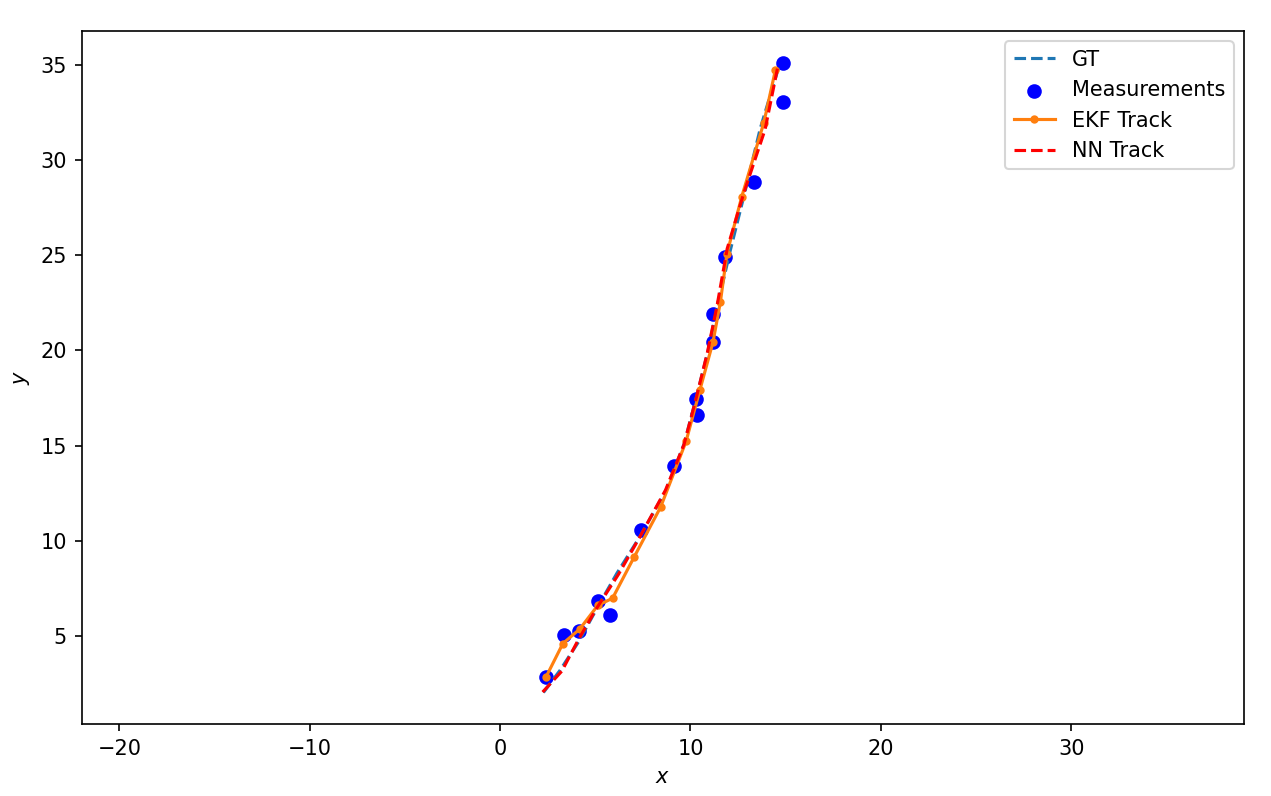}}
    \caption{NN Training Output, GT, \& EKF}
    \label{fig:my_label}
\end{figure}
\noindent The mean squared error for the training set is: 0.0078\\
The neural network predictions of the training set are very close to the ground truth values. Taking a closer look at the plot (Figure. 6).
%\\
%\begin{figure}[H]
%    \centering
%     \scalebox{0.8}{
%    \includegraphics[width=10cm]{all plots closeup covar 0.PNG}}
%    \caption{NN Training Output, GT, \& EKF (Close look)}
%    \label{fig:my_label}
%\end{figure}
\noindent The sum of the Euclidean distances between the neural network output of the training set and the ground truth is: 1.142\\
\noindent The sum of the Euclidean distances between the Extended Kalman Filter track corresponding to the training set and the ground truth is: 9.167.\\
The results of testing on similar cases, Figures 8, 9, and 10 below show testing results (tracks generated by the neural network) for two different cases and for 10 input measurement values (measurements values were not used in training).
\begin{figure}[H]
    \centering
     \scalebox{0.8}{
    \includegraphics[width=10cm]{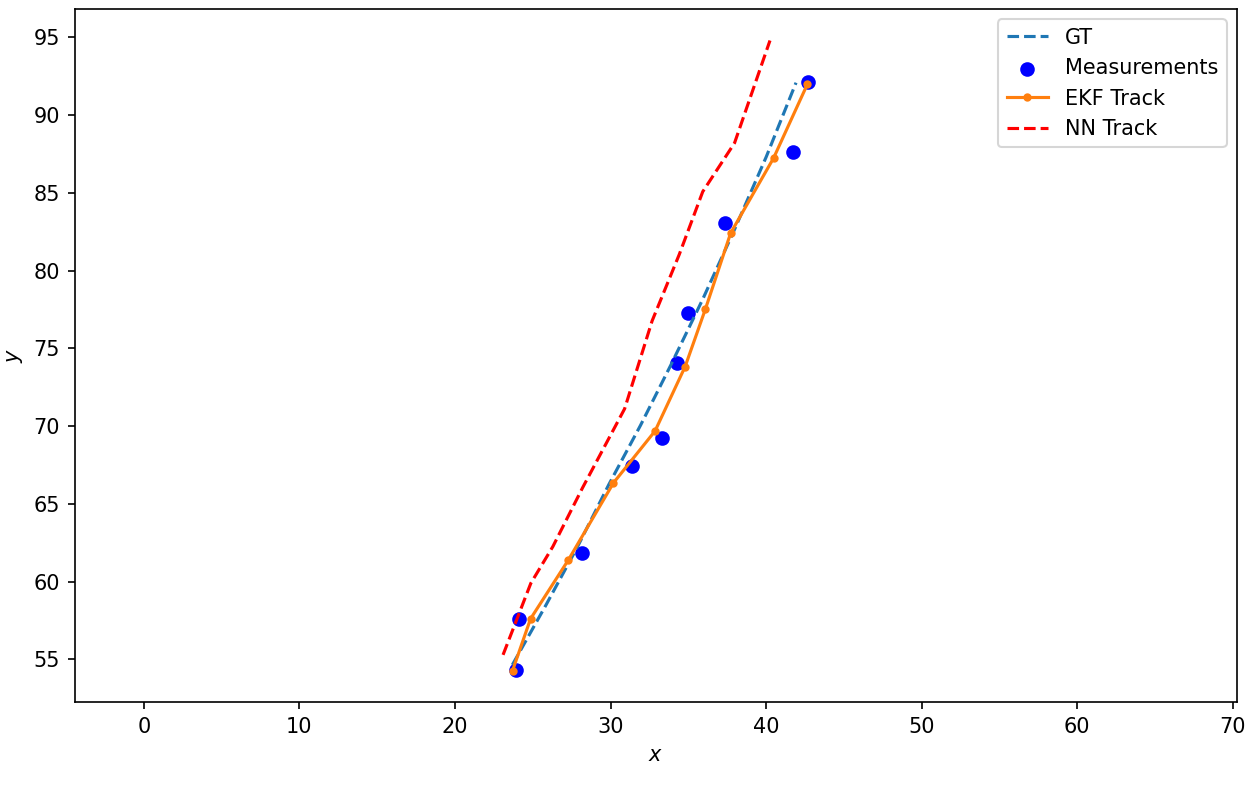}}
    \caption{Testing with Ground Truth-1 and Cov-1}
    \label{fig:my_label}
\end{figure}
\begin{figure}[H]
    \centering
     \scalebox{0.8}{
    \includegraphics[width=10cm]{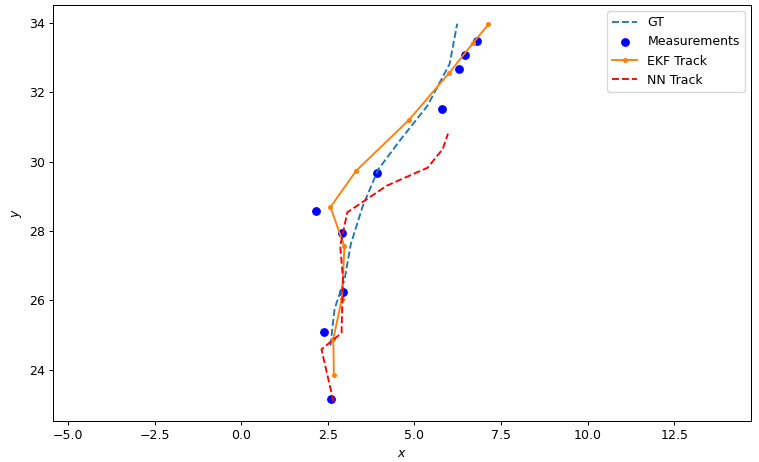}}
    \caption{Testing with Ground Truth-2 and Cov-1}
    \label{fig:my_label}
\end{figure}
\begin{figure}[H]
    \centering
     \scalebox{0.8}{
    \includegraphics[width=10cm]{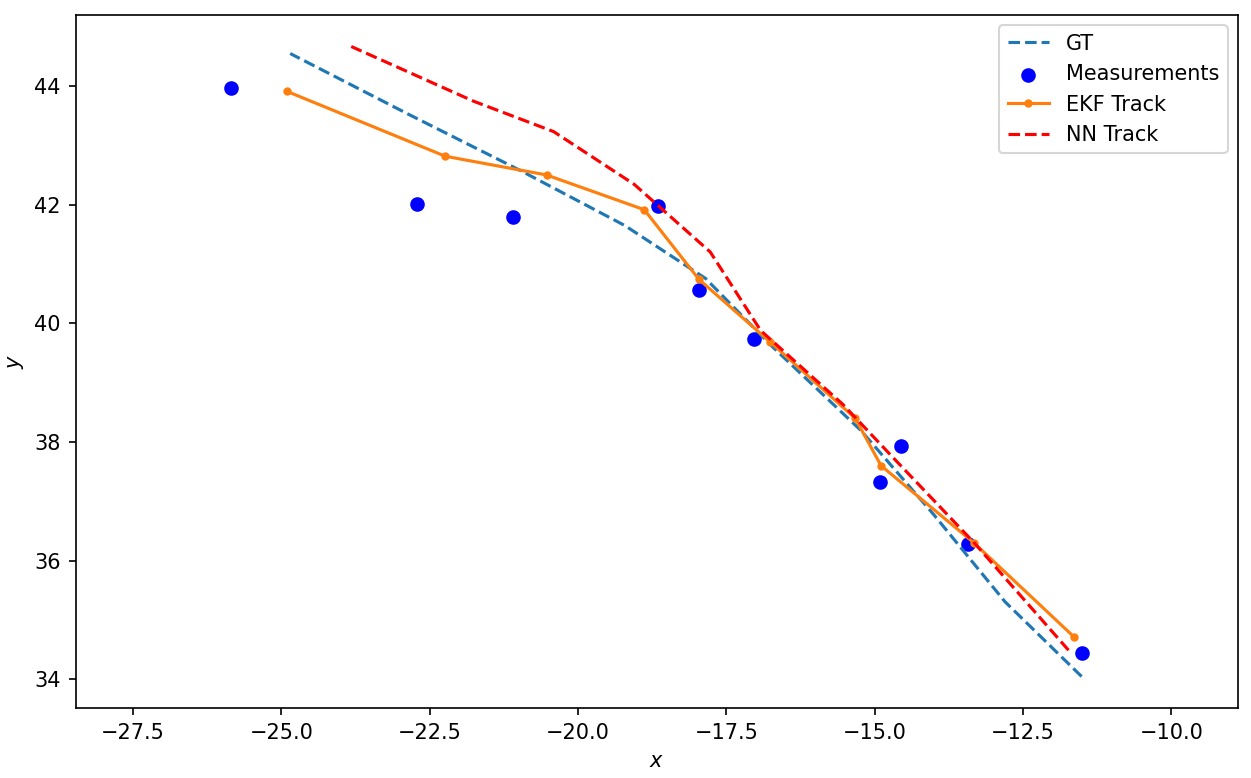}} 
    \caption{Testing with Ground Truth-4 and Cov-1}
    \label{fig:my_label}
\end{figure}

%% to here 26/9/2022
More general and comprehensive results can be summarized in Table IX and depicted in Figure 10. Results show outcomes of testing with three different ground truth tracks and five different pairs of covariance values that were used in generating the measurements values as follows: cov1=(0.01,0.3),cov2=(0.11,0.5),cov3(0.15,0.7), cov4=(0.2,1) and cov5=(0.3,1.5). The first covariance value is for the bearing angle and it is in radians and the second covariance value is for the range and it is in meters. The sum of Euclidean distances (RMS) values is calculated as a measure of quality. The RMS measures the difference between the estimated track values (generated by the Neural Network or the EKF) and the ground truth calculated over 10 successive inputs of measurements (again, those measurements values were not used in training):\\

%begin{figure}[H]
 %   \centering
 %   \includegraphics[width=10cm]{GT3 Cov5 Measurments.PNG} 
%\caption{GT3, Cov5 \& Measurements Plot}
  %  \label{fig:my_label}
%\end{figure}

%onecolumn
%begin{table}[H]
\begin{center}
\text{Table IX: Summary of Results}
\begin{figure}[H]
\scalebox{0.7}{
%\centering
    \includegraphics{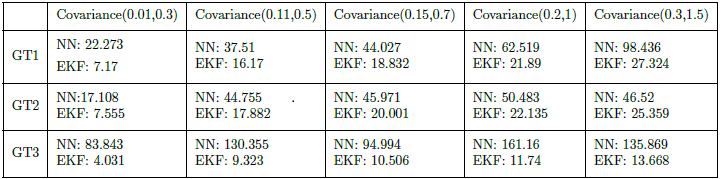}}
    %\caption{Histogram}
    %\label{fig:Table summary}
\end{figure}
\end{center}
%\scalebox{0.8}
%\begin{tabular}{|l|l|l|l|l|l|}
%\hline
%& Covariance(0.01,0.3) & %Covariance(0.11,0.5) & Covariance(0.15,0.7) & Covariance(0.2,1) & Covariance(0.3,1.5)\\
%\hline
%GT1 & \makecell[l]{NN: 22.273\\EKF: 7.17 } & \makecell[l]{NN: 37.51\\EKF: 16.17} & \makecell[l]{NN: 44.027\\EKF: 18.832} & \makecell[l]{NN: 62.519\\EKF: 21.89} & \makecell[l]{NN: 98.436\\EKF: 27.324}
%\\
%\hline
%GT2 & \makecell[l]{NN:17.108\\EKF: 7.555 } & \makecell[l]{NN: 44.755\\EKF: 17.882} & \makecell[l]{NN: 45.971\\EKF: 20.001} & \makecell[l]{NN: 50.483\\EKF: 22.135} & \makecell[l]{NN: 46.52\\EKF: 25.359}  \\
%\hline
%GT3 & \makecell[l]{NN: 83.843\\EKF: 4.031 }& \makecell[l]{NN: 130.355
%\\EKF: 9.323} & \makecell[l]{NN: 94.994\\EKF: 10.506} & \makecell[l]{NN: 161.16\\EKF: 11.74} & \makecell[l]{NN: 135.869\\EKF: 13.668}\\
%\hline
%end{tabular}
%\label{tab:figure3}
%\end

%%\onecolumn
%\vspace{4cm}
\begin{figure}[H]
    \begin{center}
    \includegraphics[width=8cm, height=8cm]{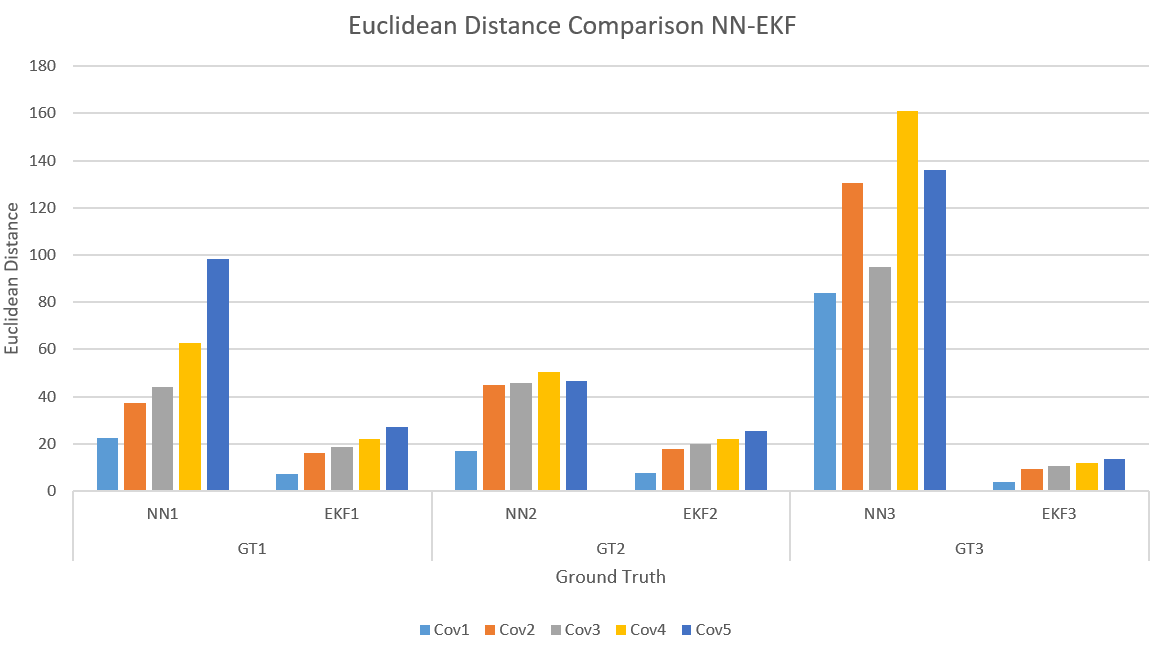}
    \caption{Histogram for the values in Table IX}
    \label{fig:my_label}
    \end{center}
\end{figure}

%\twocolumn
%\end{document}
%\section{Conclusions and Discussions}
%This work proposes a sliding window neural network for estimating the tracks of the flying objects. The measurements were simulated by stone soup software based on a Gaussian Noise model. Unlike the Kalman Filter, the proposed neural network does not use a motion model, as in some cases it is difficult to decide or select a proper motion model to use. Moreover, linearization of the motion models can be very costly. Besides, the neural network, once trained,  does not require the complex   prediction/update process that the KFs have. The training was done based on several given ground truth tracks, testing was based on new samples of measurements. Once training is finished, the inference (testing) is usually fast and efficient, it is simply a single phase of multiplications and additions. The results showed that the proposed (no motion model) system can achieve comparable results to the Extended Kalman Filter (EKF) (which is an advanced form of Kalman ). Results were close to the (EKF) results when using low measurements noise covariance values and for most of the ground, tracks used. The proposed system can be used in cases when Signal-to-Noise ratios are high and when it is difficult to find a proper motion model to use by the EKF. Future work will include incorporating motion model within the neural tracker in order to achieve better tracking accuracy and more reliable performance. 
%%===========================================================================================%%
\section{Summary}
%%%%%%%%%%%%%%%%%%%%%%%%%%%%%%%%%%%%%%%%%%%%%%%%%%%%%
This work proposes a sliding window neural network for estimating the tracks of flying objects. The measurements were simulated by stone soup software based on a Gaussian Noise model. Unlike the Kalman Filter, the proposed neural network does not use a motion model, as in some cases it is difficult to decide or select a proper motion model to use. Moreover, linearization of the motion models can be very costly. Besides, the neural network, once trained,  does not require the complex   prediction/update process that the KFs have. The training was done based on several given ground truth tracks, testing was based on new samples of measurements. Once training is finished, the inference (testing) is usually fast and efficient, it is simply a single phase of multiplications and additions. The results showed that the proposed (no motion model) system could achieve comparable results to the Extended Kalman Filter (EKF) (which is an advanced form of Kalman ). Results were close to the (EKF) results when using low measurements noise covariance values and for most ground truth tracks. The proposed system can be used in cases when Signal-to-Noise ratios are high and when it is difficult to find a proper motion model to use by the EKF. Future work will include incorporating a motion model within the neural tracker in order to achieve better tracking accuracy and more reliable performance. It is clear also that proposing another neural network to estimate the states of the object model will increase the accuracy of the estimated tracks. In this study, we used only a measurement model to estimate the tracks with the use of the neural network. Future work will include a dual neural network (one for the measurements and the other one for the object dynamics) for learning the estimation of the tracks.

%%%%%%%%%%%%%%%%%%%%%%%%%%%%%%%%%%%%%%%%%%%%%%%%%%%%%%%%%%%%%%%%%%%%%%%%%%%%%%%%%%%%%%%%%%%%%%%%%
%\appendices{}              % note there is no {} to put a title. Each appendix has its own title
%%%%%%%%%%%%%%%%%%%%%%%%%%%%%%%%%%%%%%%%%%%%%%%%%%%%%%%%%%%%%%%%%%%%%%%%%%%%%%%%%%%%%%%%%%%%%%%%%
% For a single appendix, use the \appendix{} keyword and do not use the \section command.

%\section{More Information}        % first appendix
%%%%%%%%%%%%%%%%%%%%%%%%%%
%This is the first appendix. 

%\subsection{Comments}
%If you have only one appendix, use the ``appendix'' keyword.

%\subsection{More Comments}
%Use section and subsection keywords as usual.

%\section{Yet More Information}    % second appendix
%%%%%%%%%%%%%%%%%%%%%%%%%%%%%%
%This is the second appendix.

%%%%%%%%%%%%%%%%%%%%%%%%%%%%%%%%%%%%%%%%%%%%%%%%%%%%%%%%%%%%%%%%%%%%%%%%%%%%%%%%%%%%%%%%%%%%%%%%%%%%%%
\acknowledgments
This work was partially Funded by the Thales Chair of Excellence project, Sorbonne Center of Artificial Intelligence, Sorbonne University, Abu Dhabi, UAE.

%%%%%%%%%%%%%%%%%%%%%%%%%%%%%%%%%%%%%%%%%%%%%%%%%%%%%%%%%%%%%%%%%%%%%%%%%%%%%%%%%%%%%%%%%%%%%%%%%%%%%%
\bibliographystyle{IEEEtran}
\bibliography{main}
%\bibliography{IEEEabr,MyBibFile}
%\begin{thebibliography}{1}

%\bibitem{ITAR}
%U.S. Munitions List, Sections 38 and 47(7) of the %Arms Export Control Act (22 U.S.C 2778 and 2794(7).

%\bibitem{AeroConf}
%Conference Web site: \underline{www.aeroconf.org}.

%\end{thebibliography}

%%%%%%%%%%%%%%%%%%%%%%%%%%%%%%%%%%%%%%%%%%%%%%%%%%%%%%%%%%%%%%%%%%%%%%%%%%%%%%%%%%%%%%%%%%%%%%%%%%%%%%
\thebiography
%% This biostyle allows you to insert your photo size 1in X 1.25in
\begin{biographywithpic}
{Haya Ejjawi}{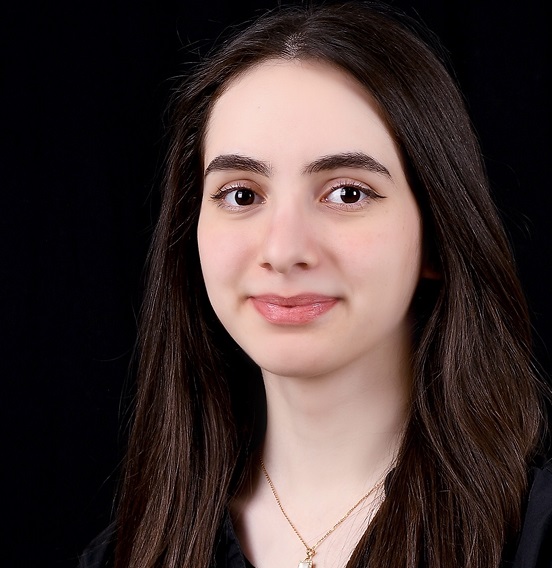}
is a student at the department of Physics and Sciences, Sorbonne University, Abu Dhabi, and was an intern with Thales/SCAI.  Her interests are in machine learning and drone recognition and tracking.
\end{biographywithpic}
\vspace{1.5cm}
\begin{biographywithpic}
{Amal El Fallah Seghrouchni}{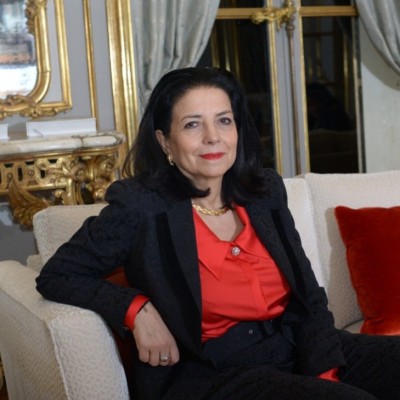}
is the Head of Ai Movement, International AI Center of Morocco, UM6P, Full Professor@Sorbonne Univ, Paris, France, LIP6 and CNRS, Member of the COMEST, UNESCO - and General Chair of AAMAS 2020 (Auckland - NZ). She is the author of more than 100 publications and supervised more than 33 Ph.D. students. Her research interests are in multi-agent systems, AI systems, and ambient intelligence. 
\end{biographywithpic} 
%\vspace{1cm}
\begin{biographywithpic}
{Frederic Barbaresco}{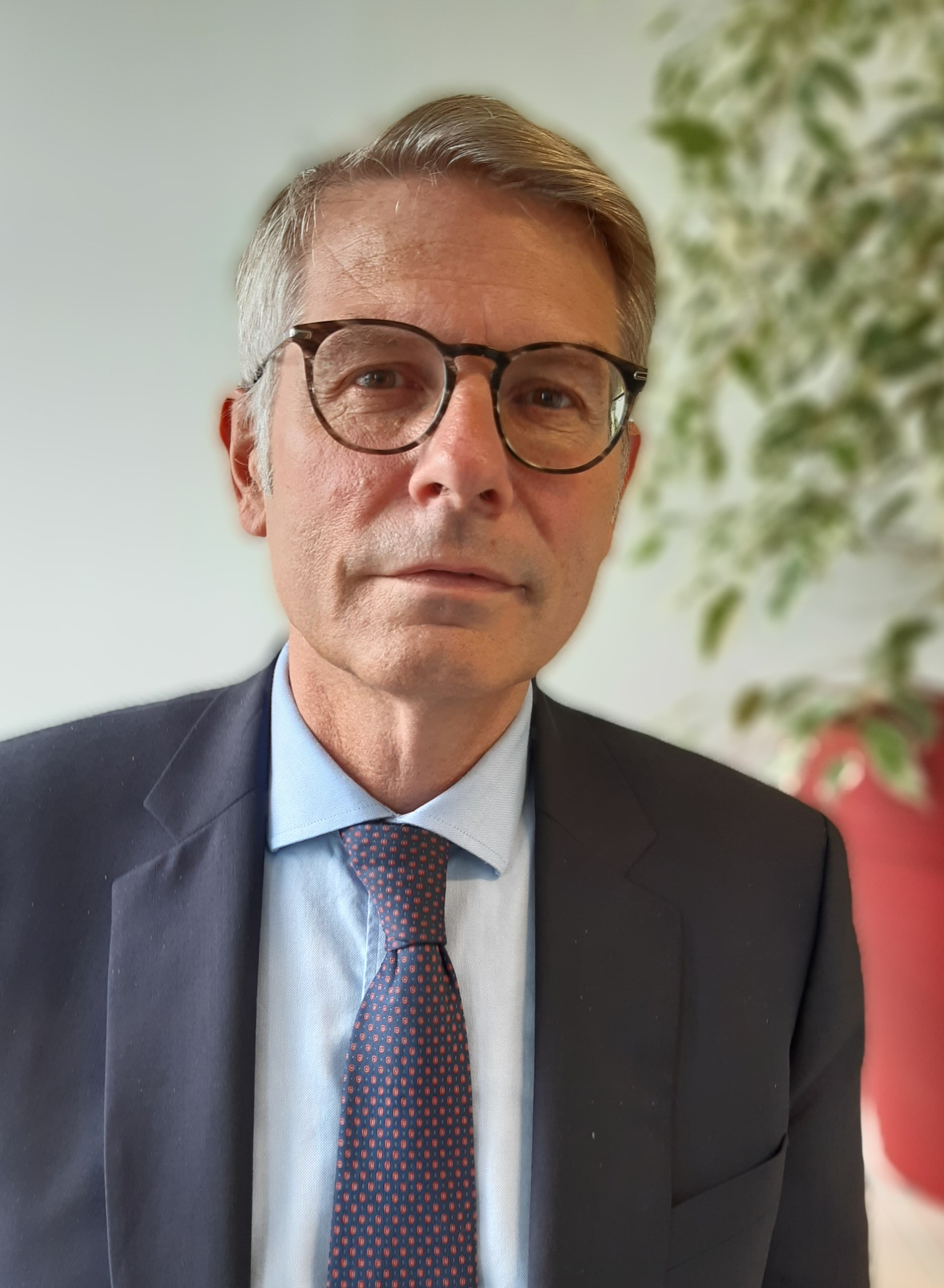}
is a senior THALES Expert in Artificial Intelligence at the Technical Department of THALES Land and Air Systems. SMART SENSORS Segment Leader for the THALES Corporate Technical Department (Key Technology Domain "Processing, Control, and Cognition"). THALES representative at the AI Expert Group of ASD (AeroSpace and Defense Industries Association of Europe). 2014 Aymée Poirson Prize of the French Academy of Science for the application of science to industry. Ampère Medal, Emeritus Member of the SEE, and President of the SEE ISIC club "Information and Communication Systems Engineering". He is a French MC representative of European COST Calista. General Chair of several elite and highly specialized conferences. Guest Editors of Special Issues "Lie Group Machine Learning and Lie Group Structure Preserving Integrators". Author of more than 200 scientific publications and more than 20 patents.
\end{biographywithpic} 

\begin{biographywithpic}
{Raed Abu Zitar}{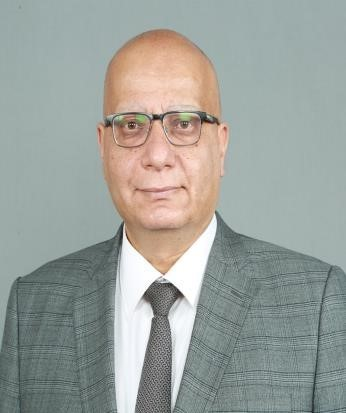}
obtained his Ph.D. in computer engineering from Wayne State University, USA, in 1993. He is an IEEE member and a senior research scientist in the context of the Thales Endowed Chair of Excellence, SCAI, Sorbonne University, Abu Dhabi, UAE. He has been with SCAI/Sorbonne University, Abu Dhabi, since Feb-2021. 
He  has more than 75 publications and is engaged in many funded projects along with graduate students' supervision. His research interests are in machine learning, neural networks, cyber security, and stochastic processing.  
\end{biographywithpic}

%\end{document}

%\bibliographystyle{elsarticle-num-names}
%\bibliography{reference}

\end{document}